\RequirePackage{fix-cm}
\documentclass[onecolumn]{svjour3}       
\usepackage{graphicx,longtable,booktabs,siunitx}
\usepackage{multicol}
\usepackage{supertabular}
\smartqed  
\usepackage{appendix}
\usepackage{amsmath}
\usepackage{graphicx}
\usepackage{lineno}
\usepackage{array}
\usepackage{longtable}
\usepackage{multirow}
\usepackage{adjustbox}
\usepackage{color}
\usepackage{afterpage}
\usepackage{algpseudocode}
\usepackage[ruled, vlined, linesnumbered]{algorithm}
\usepackage{array}
\usepackage{makecell}

\newcommand{\Pexp}{\mathbin{\text{$\vcenter{\hbox{\textcircled{$e$}}}$}}}
\usepackage{lscape}
\usepackage[square,numbers,super]{natbib}

\newcommand*\patchAmsMathEnvironmentForLineno[1]{%
\expandafter\let\csname old#1\expandafter\endcsname\csname #1\endcsname
\expandafter\let\csname oldend#1\expandafter\endcsname\csname end#1\endcsname
\renewenvironment{#1}%
{\linenomath\csname old#1\endcsname}%
{\csname oldend#1\endcsname\endlinenomath}}%
\newcommand*\patchBothAmsMathEnvironmentsForLineno[1]{%
\patchAmsMathEnvironmentForLineno{#1}%
\patchAmsMathEnvironmentForLineno{#1*}}%
\AtBeginDocument{%
\patchBothAmsMathEnvironmentsForLineno{equation}%
\patchBothAmsMathEnvironmentsForLineno{align}%
\patchBothAmsMathEnvironmentsForLineno{flalign}%
\patchBothAmsMathEnvironmentsForLineno{alignat}%
\patchBothAmsMathEnvironmentsForLineno{gather}%
\patchBothAmsMathEnvironmentsForLineno{multline}%
}

\begin{document}

\title{Multi-task Recurrent Neural Networks to Simultaneously Infer Mode and Purpose in GPS Trajectories
}


\author{Ali~Yazdizadeh \and
        Arash Kalatian \and
        Zachary Patterson \and
        Bilal Farooq
}


\institute{ First and Fourth Authors\at
              Laboratory of Innovations in Transportation, Ryerson University, Canada 
              \email{ali.yazdizadeh@ryerson.ca,
              bilal.farooq@ryerson.ca}           
           \and
           Second Author  \at
               Institute for Transport Studies (ITS), University of Leeds, Uk  \email{a.kalatian@leeds.ac.uk} 
           \and
           Third Author \at
               Institute for Information Systems Engineering, Concordia University, Canada.
              \email{zachary.patterson@concordia.ca}
}

\date{Received: DD Month YEAR / Accepted: DD Month YEAR}

\maketitle

\begin{abstract}
Multi-task learning is assumed as a powerful inference method, specifically, where there is a considerable correlation between multiple tasks, predicting them in an unique framework may enhance prediction results. This research challenges this assumption by developing several single-task models to compare their results against multi-task learners to infer mode and purpose of trip from smartphone travel survey data collected as part of a smartphone-based travel survey. GPS trajectory data along with socio-demographics and destination-related characteristics are fed into a multi-input neural network framework to predict two outputs; mode and purpose. We deployed Recurrent Neural Networks (RNN) that are fed by sequential GPS trajectories. To process the socio-demographics and destination-related characteristics, another neural network, with different embedding and dense layers is used in parallel with RNN layers in a multi-input multi-output framework. The results are compared against the single-task learners that classify mode and purpose independently. We also investigate different RNN approaches such as Long-Short Term Memory (LSTM), Gated Recurrent Units (GRU) and Bi-directional Gated Recurrent Units (Bi-GRU). The best multi-task learner was a Bi-GRU model able to classify mode and purpose with an F1-measures of 84.33\% and 78.28\%, while the best single-task learner to infer mode of transport was a GRU model that achieved an F1-measure of 86.50\%, and the best single-task Bi-GRU purpose detection model that reached an F1-measure of 77.38\%. While there's an assumption of higher performance of multi-task over sing-task learners, the results of this study does not hold such an assumption and shows, in the context of mode and trip purpose inference from GPS trajectory data, a multi-task learning approach does not bring any considerable advantage over single-task learners.

\end{abstract}

\section{Introduction}\label{intro}
During the last decade, transportation practitioners and planners have begun using smartphone-based surveys to collect  respondent trajectory and other trip-related information. Researchers have investigated various inference methods to elicit  trip information from smartphone-based travel surveys, especially to detect mode of transport and trip purpose. 

In recent years, deep neural networks have demonstrated remarkable accomplishments in various fields of science, particularly in natural language processing and image recognition tasks. Concerning GPS trajectory analysis, deep learning approaches allow discovering factors related to each GPS point that are overlooked by traditional machine learning algorithms that use aggregated features for whole trips. 


Deep learning models are typically applied to predict a single task by training one model or an ensemble of models. However, by focusing on training a single-task model we may ignore advantageous representations derived from shared layers of related tasks. Taking into account the correlation between related tasks may improve the prediction accuracy of deep learning models. In the deep learning literature, this approach is called Multi-Task Learning \cite{ruder2017overview}.  

Other names have been used in the literature to refer to "Multi-task learning": learning with auxiliary information or tasks, joint learning, or learning to learn \cite{ruder2017overview}. 

Ultimately, any deep learning model dealing with the optimization of two or more loss functions can be considered as a multi-task learner. Even supposing our goal is to optimize a single loss function, as is the case in many neural network implementations, we can benefit from auxiliary information or tasks to improve our main task 

\cite{ruder2017overview,caruana1998multitask}.
As Caruana \cite{caruana1998multitask} has summarized the goal of multi-task learning: ``Multi-task learning improves generalization by leveraging domain-specific information contained in the training signals of related tasks.'' 

In transportation demand analysis, trip purpose and mode of transport are arguably the most important trip characteristics. Importantly, mode of transport and trip purpose are highly correlated \cite{mokhtarian2001derived}. For example, due to the limited number of parking spaces in Central Business Districts (CBD) in metropolitan areas, many travelers destined there use public transit for commuting. Hence, we can say work trips have high correlation with public transit, as mode of transport and the trip purpose, work. As a result, transportation researchers in the past have investigated joint discrete choice modeling approaches to model mode and destination choice together \cite{newman2010hierarchical}. However, multi-task (joint) machine/deep learning approaches to infer mode and trip purpose from travel surveys have not been studied in the literature. Therefore, due to the correlation between mode and purpose, as well as the joint role of mode and purpose in transportation demand, this study attempts to infer mode and purpose simultaneously from trajectory data using deep learning models.
Indeed, the study challenges the assumption of multi-task learners, that inferring two (or more) tasks simultaneously may improve the overall model performance over single-task learners. In our specific case, the results demonstrates whether there's a benefit in multi-task over single-task learners to infer mode and trips purpose from travelers' trajectory data.  

Consequently, we feed point-wise trajectory information as input into different multi-task Recurrent Neural Networks (RNN) architectures to infer mode of transport and trip purpose. As well, the models take advantage of single-observation auxiliary data, such as socio-demographics, to improve the prediction accuracy of mode and purpose of trips.

The study has been conducted using data collected by the smartphone travel survey app, MTL Trajet, which is an instance of the smartphone travel survey platform, Itinerum \cite{patterson2019itinerum}. The MTL Trajet data was collected as part of a large-scale pilot study on the 17\textsuperscript{th} of October 2016 in a 30-day travel survey study, in which over 8,000 respondents participated \cite{patterson2017mtl}.

The rest of the paper is organized as follows: a background section describes previous work on multi-task learning and RNNs in mode or activity detection from trajectory data. The methodology section sets out the framework of the RNN models as well as of the data description and pre-processing. The next section after that presents the results of the different multi-task RNN architectures on the MTL Trajet dataset. The last section concludes the paper.

\section{Background}
This section reviews previous research related to mode and purpose detection from smartphone data and briefly introduces RNN models, especially Long Short-Term Memory (LSTM) and Gated Recurrent Units (GRU). Comprehensive reviews on mode and purpose detection have been conducted by Wu et al.~\cite{wu2016travel}, Elhoushi et al.~\cite{elhoushi2017survey}, and Wang et al.~\cite{wang2018applying} so we keep ours short here. Similarly, different Recurrent Neural Network approaches have been explained in detail by Goodfellow et al. \cite{goodfellow2016deep}. Also, recent applications of multi-task learning have been fully covered by Ruder et al. \cite{ruder2017overview}.

Mode and purpose detection has been studied in the literature using various approaches, which include: rule-based, machine learning, deep learning, and discrete choice approaches. Ensemble tree-based methods \cite{xiao2017identifying}, Random Forest models \cite{yazdizadeh2018automated}, hybrid rule-based and Random Forest approaches \cite{wang2018travel} and other machine learning models have been widely used in the literature to detect mode of transport. Methods from other fields of science have been applied for the mode detection task. For example, Assemi et al. \cite{assemi2016developing} utilized a nested logit model to infer mode of transport from smartphone travel surveys. The next section reviews the studies on mode and purpose detection from mobile phone data using deep learning approaches.

\subsection{Mode and Purpose Detection}
In recent years, deep neural network models, such as Convolutional Neural Networks (CNN) and Recurrent Neural Networks have been deployed for mode detection. Convolutional Neural Networks (CNN) have been used by Dabiri and Heaslip \cite{dabiri2018inferring} and Yazdizadeh et al. \cite{yazdizadeh2019ensemble} to detect the mode of transport. Endo et al. \cite{endo2016deep} suggested a framework to automatically calculate trajectory features by converting GPS-points along a trip into a two-dimension (2D) image structure. They then deployed a deep learning model and fed it with the ``stay time'' of each GPS point (analogous to the pixel values of an image). Finally, they used traditional machine learning models, such as logistic regression and decision trees to predict mode of transport based on pixel values and other hand-crafted features. 

Long-Short Term Memory (LSTM), Gated Recurrent Units (GRU) and control gate-based Recurrent Neural Networks (CGRNN) are the most well-known Recurrent Neural Networks methods to analyse the sequential data and have been applied on GPS trajectories to detect the mode of transport \cite{vu2016transportation,otte2012local}. Vu et al.~\cite{vu2016transportation} analyzed different RNN approaches and found the superiority of GRU and CGRNN models over simple RNN and LSTM models to infer mode of transport from accelerometer data. Simoncini et al. \cite{simoncini2018vehicle} used an RNN model for vehicle classification from low-frequency GPS data\footnote{Vehicle classification studies focus on determining the type of vehicle, for example categorizing vehicles based on the number of axles.}. They used different features, such as timestamp, longitude and latitude, speed and odometer to detect vehicle type. They used an RNN architecture with a 1-D pooling layer that aggregates the final recurrent layer of RNN, to predict vehicle type.  

Liu et al.~\cite{liu2017end} developed a bidirectional LSTM (Bi-LSTM) to classify mode of transport from GPS data. The Bi-LSTM model demonstrated good performance while only considering time interval, longitude, and latitude of trajectory GPS data.  Moreover, they suggested an enhanced embedding Bi-LSTM model that considers time interval as an external feature fed into an embedding layer, where the Bi-LSTM part uses the same architecture as their base Bi-LSTM model.

With respect to purpose detection, there are fewer studies in the literature compared to mode inference studies. While mode detection is related to GPS-point features such as speed and acceleration, trip purpose has more correlation with trip destination and traveler attributes, such as different land-use types in the vicinity of a destination, socio-demographics, and other trip attributes such as time-of-day or day-of-week of the \cite{yazdizadeh2018automated}. While trip trajectories can be treated as sequential observations and fed to RNN or CNN models to detect the mode of transport, trip destination is a single-point observation, and cannot be fed to RNN or CNN models. This limitation has caused researchers to use more classical machine learning algorithms, such as Random Forest \cite{yazdizadeh2018automated,montini2014trip}, or rule-based methods \cite{fang2018identifying} for purpose detection. However, due to the correlation between mode and purpose in travel choice behavior, one may expect better prediction accuracies while training a single multi-task model, rather than training two single-task learners. Joint mode and purpose discrete choice modeling has been investigated by a few studies~\cite{newman2010hierarchical,anas1981estimation} using household travel data. Newman and Bernardin \cite{newman2010hierarchical} explored the relationship between mode and purpose choice in an integrated nested choice model. A Multinomial Logit model also was developed \cite{anas1981estimation} to estimate the joint location and travel mode choice using aggregated travel data. However, joint mode and purpose inference from trajectory data using deep learning approaches has not been well covered in the literature. At the same time, multi-task learning is known as a powerful inference method in machine learning \cite{augenstein2018multi}. Multi-task learning has been used in traffic data imputation studies \cite{rodrigues2018multi}. Rodrigues et al.~\cite{rodrigues2018multi} proposed multi-output Gaussian processes (GPs) to model spatial and temporal patterns in traffic data. They demonstrated that a multi-output GP model was able to capture the complex dependencies and correlations between nearby road segments to improve imputation accuracy. Their multi-output model outperforms other imputation methods, such as a Recurrent Neural Network model, by taking into account the complex
spatial dependencies between subsequent road segments. 

As explained above, while single-task inference models have been studied on trajectory data to infer mode and purpose, the current literature lacks studies on multi-task mode and purpose inference. Hence, the present study adopts a multi-task approach, i.e. a multi-task learner, based on both trip trajectory and destination characteristics, and socio-demographic information. By implementing a multi-task learning approach, we aim to capture and exploit the correlation between mode of transport and trip purpose in our dataset and find out to what extent such an approach can enhance the prediction performance for both mode and purpose of trips. The model architecture is explained in methodology section. 

\subsection{Recurrent Models}
Recurrent Neural Networks have been a popular choice for training sequential data \cite{goodfellow2016deep}. However, their capability is restricted by a lack of memorization of long and short term dependencies across sequential data points, due to the vanishing or exploding of gradients in long sequences \cite{bengio1994learning}. LSTM and GRU approaches were developed to overcome these shortcomings of RNN models. We briefly explain each component of LSTM and GRU models and their differences.\\

\subsubsection{Inside an RNN Block}
In the core of each RNN model, there is a recurrent hidden state described by \cite{dey2017gate}:

\begin{equation}\label{equ:hidden state}
    h_t = g(Wx_t + Uh_{t-1} + b)
\end{equation}

where $x_t$ is the input vector (in our case a vector of GPS points along a trip) at time $t$, $h_t$ is the hidden state at time $t$, $g$ is an activation function, $W$ and $U$ are trainable weights, and $b$ is a trainable bias. Finally, $h_{t-1}$ is the hidden state (or output) of the previous time step $t-1$. Hence, each RNN block has two inputs: $x_{t}$ and $h_{t-1}$, and outputs the hidden state at the current time step, i.e. $h_t$ to the next RNN block.

\subsubsection{Inside an LSTM Block}
The LSTM model introduces three gating (control) signals: \textit{input}, \textit{forget} and \textit{output} gating signals at each time step $t$. These gating signals are similar to Equation~\ref{equ:hidden state}, with weights and bias parameters, and a sigmoid activation function. One can imagine these gating signals as valves that control the flow of water in pipelines. However, these valves, with sigmoid functions, control the flow of memory, input, and output between LSTM blocks.  

The \textit{forget, input} and \textit{output} gating signals have the following equations:

\begin{equation}\label{equ:forget signal_lstm}
    f_t = \sigma(W_f x_t + U_f h_{t-1} + b_f)
\end{equation}
\begin{equation}\label{equ:input signal_lstm}
    i_t = \sigma(W_i x_t + U_i h_{t-1} + b_i)
\end{equation}
\begin{equation}\label{equ:output signal_lstm}
    o_t = \sigma(W_o x_t + U_o h_{t-1} + b_o)
\end{equation}
where $\sigma$ is a sigmoid function. 

Moreover, LSTMs benefit from using internal memory cells, $\widetilde{c}_{t}$, inside each LSTM block. The flow of memory between blocks is determined by the \textit{forget} and \textit{input} control signals. Indeed, an LSTM block can forget the memory, when the output of the sigmoid function of a \textit{forget} gating signal is 0, or keep the memory if the sigmoid outputs 1. 

As shown in Equation~\ref{equ:forget signal_lstm}, the \textit{forget} gating signal $f_t$ is calculated based on the hidden state of the previous time step $h_{t-1}$, and the input of the current LSTM block $x_t$. The output of the sigmoid function of the \textit{forget} gating signal $f_t$ is then applied on the old memory cell $c_{t-1}$ by a element-wise multiplication, as in Equation~\ref{equ:new memory cell_lstm}:

\begin{equation}\label{equ:new memory cell_lstm}
    c_t = f_t \Pexp c_{t-1} + i_t \Pexp \widetilde{c}_{t}
\end{equation}

The element-wise multiplication is denoted by $\Pexp$. The second term in the above equation controls how much of the memory in the current memory cell $\widetilde{c}_{t}$ should influence the new memory, i.e. $c_t$. This role is played by the \textit{input} gating signal while multiplied (element-wise) with the $\widetilde{c}_{t}$. The current memory cell $\widetilde{c}_{t}$ is defined as:

\begin{equation}\label{equ:c tilda_lstm}
    \widetilde{c}_{t} = g(W_c x_t + U_c h_{t-1} + b_c)
\end{equation}
where $g$ is usually the hyperbolic tangent function or the rectified Linear Unit (ReLU) \cite{dey2017gate}. 

After calculating the current memory cell of an LSTM block with Equations~\ref{equ:forget signal_lstm}, \ref{equ:input signal_lstm}, \ref{equ:new memory cell_lstm} and \ref{equ:c tilda_lstm}, the hidden state, $h_t$ of the LSTM block is obtained by the following formula:

\begin{equation}\label{equ: h_lstm}
    h_{t} = o_t \Pexp g(c_t)
\end{equation}

where the $o_t$ is the \textit{output} gating signal, defined by Equation~\ref{equ:output signal_lstm}, which controls how much memory should transfer to the next LSTM block. 

As explained in the above equations, the distinctive characteristic of the LSTM model is the concept of the memory cell, which is passed through LSTM blocks and the flow of memory is regulated by the implementation of different gating signals. \\

\subsubsection{Inside a GRU Block}
The main difference between an LSTM and GRU block is the number of gating signals, with two gating signals in GRU instead of three gating signals in LSTM. GRUs possess an update gate, denoted by $z_t$, and a reset gate, $r_t$ \cite{dey2017gate}. The hidden state of a GRU block is calculated based on the following equations \cite{dey2017gate}:

\begin{equation}\label{equ:c tilda}
    h_t = (1-z_t) \Pexp h_{t-1} + z_t \Pexp \widetilde{h}_t
\end{equation}

\begin{equation}\label{equ:c tilda_2}
    \widetilde{h}_t = g(W_h x_t + U_h(r_t \Pexp h_{t-1} + b_h)
\end{equation}

The two gating signals in GRU blocks are defined by similar formulas for the gating signals in an LSTM:

\begin{equation}\label{equ:input signal}
    z_t = \sigma(W_z x_t + U_z h_{t-1} + b_z)
\end{equation}
\begin{equation}\label{equ:output signal}
    r_t = \sigma(W_r x_t + U_r h_{t-1} + b_r)
\end{equation}

Various studies have demonstrated the comparable performance of GRU and LSTM in many cases \cite{chung2014empirical}, although some studies have found that GRU outperforms LSTM models \cite{dey2017gate,chung2014empirical}. 

\section{Methodology}\label{method}
In this section, we describe the data used, data pre-processing, the RNN architectures used and hyperparameter value determination. 

\subsection{Data Used}\label{data_prep}
The MTL Trajet dataset consists of three types of data; validated mode and purpose of trips, GPS coordinates, \textit{timestamps}, and respondent socio-demographic information. The data has been collected via MTL Trajet app, which is supported by an in-app mode and purpose validation process. The MTL Trajet app prompted and asked respondents to validate their mode of transport and trip purpose upon detecting a stop during their movements (stops are detected by a rule-based algorithm that works in the background of MTL Trajet app). Furthermore, upon the installation of MTL Trajet, a series of questions were asked of respondents to gather their socio-demographics, such as age, sex, occupation, home location, work or school location, etc. During the MTL Trajet 2016 survey, over 33 million GPS points were recorded. There were four modes of transport, i.e. walk, bike, car, public transit, validated by respondents. Also, six trip purposes were collected including education, health, return home, shopping, work and return home.

To detect trips, we used a rule-based trip-breaking algorithm explained in Patterson \& Fitzsimmons \cite{patterson2016datamobile}. The algorithm detects start and end point of trips based on the 3-minute dwell time between GPS points as the most prominent criterion for detecting trips. Afterward, the algorithm verifies the velocity and parameters relating to the public transit network (i.e. transit junctions and metro station location) and then stitches segments back together into complete trips. As an example, wherever two consecutive GPS points are located within 300 meters of two different metro stations (due to the sparsity of data collection when underground), and the time interval between them is less than the maximum travel time by metro, the segments are considered as part of the same trip. Also, wherever two consecutive GPS points fall within the same intersection with bus correspondences, the algorithm allows for a wider time interval (10-minute instead of 3-minute gap).

\subsection{Data Preparation}
The dataset consists of two types of data: sequential data (trip trajectories), that comes from the GPS points, and auxiliary data, i.e. single-observation data including socio-demographics and trip destination characteristics. Each trip in our dataset comprises GPS points with three GPS features: time interval, longitude, and latitude, used in the sequence part of our models. While the number of GPS points along a trip varies between 15 to more than 1000 points, we considered only the 70 points of each trip (as 70 is the average number of points along all trips in our dataset). Also, to select the 70 points for each trip, we considered the first 35 and the last 35 GPS points along each trip. When the number of points along a trip was less than 70, we padded the trajectories.

Since not all trips were validated for both mode and purpose, we are left with 7,763 trips in our dataset. The auxiliary features used are shown in Table~\ref{auxiliary feat}. Socio-demographics and trip characteristics are obtained from the MTL Trajet dataset. Land-use data are derived from Montreal land-use data (compiled by the provincial government ministry ``MAMROT'' \cite{cmm2011montreal}). The land-use categories and their frequencies have been explained in \cite{yazdizadeh2018automated}. Foursquare data was also used. It consisted of \mbox{\textit{checkinsCount}} and \mbox{\textit{usersCount}}, where \textit{checkinsCount} is the total number of check-ins in a venue near a trip destination, and \textit{usersCount} accounts for the total number of users who have ever checked into a venue near a trip destination (in a 250-meter vicinity). 

\begin{table*}[t]
\centering
\caption{Auxiliary Features Used in Mode and Purpose Detection.}
\label{auxiliary feat}
\resizebox{\textwidth}{!}{
\begin{tabular}{lll}
\cline{1-2} \hline
Attribute         & Definition     & Embedding size                                                                                  \\ \cline{1-2} \hline

\multicolumn{2}{l}{\textit{Trip Characteristics}}                                                                    \\
DAY of WEEK               & 1-7 for Monday through Sunday & 4                                                           \\
HOUR\_START             & Trip start hour, ranging from 0 to 24 &  12                                                                         \\
HOUR\_End             & Trip end hour, ranging from 0 to 24 &  12                                                                         \\

CBD\_ORIGIN       & 1: if the origin is located in Montreal's CBD, 0: otherwise       &1                               \\
CBD\_DESTIN       & 1: if the destination is located in Montreal's CBD, 0: otherwise       &1                           \\
HOME\_DEST        & Direct distance between trip destination and individual home location   &-                         \\
STUDY\_DEST       & Direct distance between trip destination and individual education location  &-                     \\
WORK\_DEST        & Direct distance between trip destination and individual work location            &-                \\
HOME\_ORG         & Direct distance between trip origin and individual home location               &-                  \\
STUDY\_ORG        & Direct distance between trip origin and individual education location          &-                 \\
WORK\_ORG         & Direct distance between trip origin and individual work location            &-                \\

MTL\_ORIGIN       & 1: if the origin is located on the Island of Montreal Island, 0: otherwise                     &1                 \\
MTL\_DESTIN       & 1: if the destination is located on the Island of Montreal Island, 0: otherwise            &1                     \\                       

\multicolumn{1}{l}{}                                                                                        \\
\multicolumn{2}{l}{\textit{Trip Destination Features}} \\
\multicolumn{2}{l}{{Land-use (number of land-use parcels in 250 meters around a trip destination)}}           \\
LU\_*   & \begin{tabular}[c]{@{}l@{}}23 different attributes each one shows the frequency of the corresponding\\ land-use category 
\end{tabular}             &-                                                                       \\
\multicolumn{1}{l}{}          &                                                                                                   \\
 \multicolumn{2}{l}{{Foursquare \textit{checkinCounts} (number of checkinCounts in 250 meters around a trip destination)}}             \\
CH\_*     & \begin{tabular}[c]{@{}l@{}}10 different attributes each one shows the checkinCounts for the corresponding\\ Foursquare category \end{tabular}     &-          \\
\multicolumn{1}{l}{}                                                                                                     \\
\multicolumn{2}{l}{{Foursquare \textit{userCounts} (number of usersCounts in 250 meters around a trip destination)}}               \\
UC\_*     & \begin{tabular}[c]{@{}l@{}}10 different attributes each one shows the usersCounts for the corresponding\\ Foursquare category \end{tabular} &- \\
\\
\multicolumn{2}{l}{\textit{Socio-demographics}}                                                                      \\


AVG\_PRICE\_NEIGH & The average value of residential buildings around each individual's home &-  \\
SEX               & 0:male, 1: female, 2: other/neither    &2                                                           \\
OCCUPATION        & \makecell[l]{0:full-time worker,  1:  part-time worker, 2: Student,\\ 3: Student and worker, 4: Retired 5: At home}  &2 \\

AGE               & 0: age between 16-24 ,1: 25-34 , 2: 35-44, 3: 45-54, 4: 55-65, 5: 65+   & 3                         \\

                                                                         \\ \cline{1-2} \hline
\end{tabular}
}
\end{table*}

\subsection{Time Transformation}
Regarding GPS features, we treat the recorded time of each GPS point as the number of seconds after midnight, which is inherently a cyclical feature. For example, 5 minutes after and before midnight, are recorded as 300 and 86,100 seconds, respectively. While the time interval between these two times is only 600 seconds, their values in seconds suggests a too-large time interval, i.e. 85,800 seconds, which does not represent the cyclical behavior of time during a day. To let the neural network recognize that an attribute is cyclical, one can transform the time into two dimensions, using the cosine and sine functions, as follows:

\begin{equation}
    sinusoid\_time = sin(2*\pi*seconds/total\_seconds)
\end{equation}
\begin{equation}
    cosinusoidal\_time = cos(2*\pi*seconds/total\_seconds)
\end{equation}

where \textit{seconds} is the number of seconds after midnight, and \textit{total\_seconds} is the total number of seconds of each day. For example, 5 minutes after midnight will represent as 0.021 in the sinusoidal dimension, and 0.999 for the cosinusoidal dimension. Also, 5 minutes before midnight will be transformed into -0.021 for the sinusoidal dimension, and 0.999 for the cosinusoidal dimension. Such transformation enables the neural network to consider the cyclical character of time over 24 hours.

\subsection{Entity Embedding}
We used the entity embedding approach \cite{guo2016entity} for the categorical data related to the trip destination or socio-demographics, instead of a one-hot Encoding approach. Both approaches have been used by researchers to deal with categorical data where there is no intrinsic ordering to the categories. The advantage of using entity embedding over  one-hot encoding is that each categorical variable is mapped into a higher dimension space which is richer in capturing the correlation between different levels of a variable. For example, "days of the week" is usually coded as a categorical variable in a 1 to 7 scale. However, with entity embedding each day of the week is usually mapped into 4-dimensional vector space enabling the model to capture the similarities, for example between "Saturday'' and "Sunday''. 

\begin{figure*}[t]
\centering
\includegraphics[keepaspectratio,width=.75\linewidth]{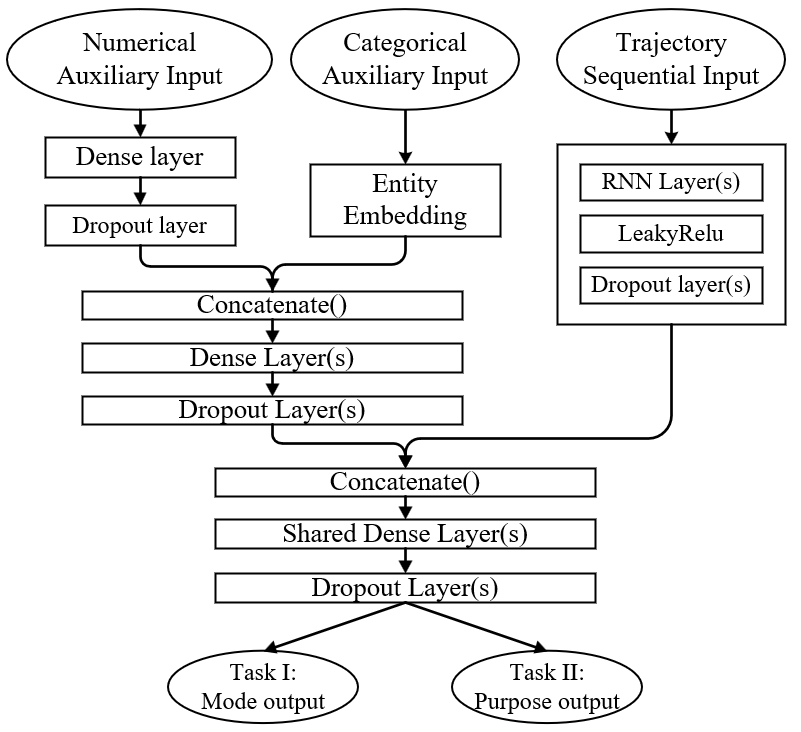}
\caption{Schematic Framework of a Multi-input Multi-output Learner to Infer Mode and Trip Purpose.
\label{fig:model architecture}}
\end{figure*}

The dataset in the current study contains 17 categorical and 50 numerical non-sequential attributes. The categorical attributes are first fed into an embedding layer and the output is then concatenated with the output of non-sequential numerical data, as shown in Figure~\ref{fig:model architecture}.

\begin{table*}[!t]

\centering
\caption{Hyperparameter values used in the Study}
\label{Table:hyper-label}
\begin{adjustbox}{width=0.6\textwidth,totalheight=0.5\textheight,keepaspectratio}
\begin{tabular}{|c|c|}
\hline
Name         & Hyperparameter value                                                                                          

\\ \hline
Recurrent layers &  \begin{tabular}[c]{@{}c@{}}Types:LSTM, GRU, Bi-LSTM, Bi-GRU,\\ Num. of layers: 1, 3, 6\\ Num. of nodes: 70, 210,\\ Activation function: Leaky ReLU \end{tabular}\\ 
\\

\\
Activity Funciton & \makecell[l]{Relu, LeakyRelu($\alpha$ = 0.05)}
\\

Dropout       & 0.2, 0.25, 0.5
\\
Output layer         & Activation: Softmax for both mode and purpose of trip                                                                                       \\
Optimization method      & \makecell[l]{Adam optimizer}                               \\
Batch size           & 16, 32, 64, 128 \\
 \hline
\end{tabular}
\end{adjustbox}
\end{table*}

\subsection{Model Architecture}
This section explains the architecture of the models to examine the performance of multi-task and single-task learners. As mentioned in the previous sections, the dataset in the current study consists of two types of data: the sequential data, which is the sequence of GPS points along a trip trajectory, and the single-observation data, which are the characteristics of the trip destination and the socio-demographics of travelers. To deal with the sequential data, we investigated different Recurrent Neural Networks architectures, developed by researchers across various fields. Among them, Long-Short Term Memory (LSTM) and Gated Recurrent Units (GRU) have gained the lion's share of attention. Also, we investigate the bi-directional GRU (Bi-GRU) to probe how the bi-directional architecture may improve the performance of the model. We also tested different hyperparameter values in Table~\ref{Table:hyper-label}.

Our approach toward the single-observation data, which contains two data types: categorical and numerical, is to develop entity embedding layers to process the categorical data, and dense layers to analyze the numerical data. A schematic of the  multi-input architecture to process all data types, i.e. trajectory sequential data, categorical auxiliary data, and numerical auxiliary is shown in Figure~\ref{fig:model architecture}. The output of the entity embedding layer and dense layers is concatenated and passed through further dense and dropout layers. Finally, the outputs from trajectory and auxiliary data are concatenated and fed into shared dense layers. 

In the multi-task learning framework, the output of the shared dense layers in Figure~\ref{fig:model architecture} is fed into two cross-entropy loss functions, for generating labels for mode and trip purpose. In single-task models, the output of the shared dense layers is fed into one loss function, for either mode or purpose classification. 

\section{Results}
This section presents the results of different single- and multi-task learners. We developed three series of models: single-task purpose classifiers, single-task mode classifiers, and multi-task mode and purpose classifiers. 
For each type of these classifiers, three different RNN architectures (i.e. LSTM, GRU, and Bi-GRU), for the RNN part of the framework were tested. The accuracy, precision, recall and F1-measure of the models is shown in Table~\ref{Table:total-results}. In multi-class classification, accuracy is defined as follows \cite{google}: \\ 
\begin{equation}
Accuracy = \frac{Correct Predictions}{Total Number of Examples}.
\end{equation}

Precision explains the frequency of the correct predicted positive class. It is defined by the following formula \cite{google}: \\ 
\begin{equation}
Precision = \frac{True Positives}{True Positives + False Positives}
\end{equation}

The F1-measure is calculated as a weighted average of the precision and recall. It is also defined by the following formula: \\ 
\begin{equation}
F1-measure= \frac{2*(Precision * Recall)}{Precision + Recall}
\end{equation}

To compare the single- and multi-task learners on comparable grounds, the characteristics of the models were consistent across all of them. The RNN part of models consists of three 70-node recurrent layers, each of which is followed by a dropout layer. The dense layers for analyzing the auxiliary numeric data in Figure~\ref{fig:model architecture} comprise one 256-node dense layer. Afterwards, the output of the final dense layer of the numerical auxiliary input, is concatenated with the output of the entity embedding layer, which processes the categorical auxiliary input. Subsequently, three 128-node dense layers process the output of concatenated layer. Afterwards, the results of RNN layers are concatenated with the results of the numerical and categorical input data. Later, the output of the three 64-node dense layers, each of which is followed by a dropout layer, amount to the shared dense layers of the framework in Figure~\ref{fig:model architecture}. We tested different dropout probability values, i.e. 0.2, 0.25 and 0.5, and found the dropout layer with a probability of 0.2 results in the best performance after the RNN layers. However, after the dense and shared layers, dropout layers with a probability of 0.5 demonstrated higher performance. While these characteristics are held constant across the multi-task and single-task learners, the output of the final layer of multi-task learners is fed into two cross-entropy loss functions. The two loss functions are then summed up, with equal weights. 

\begin{figure*}[ht]
\centering
\includegraphics[keepaspectratio,width=1.0\linewidth]{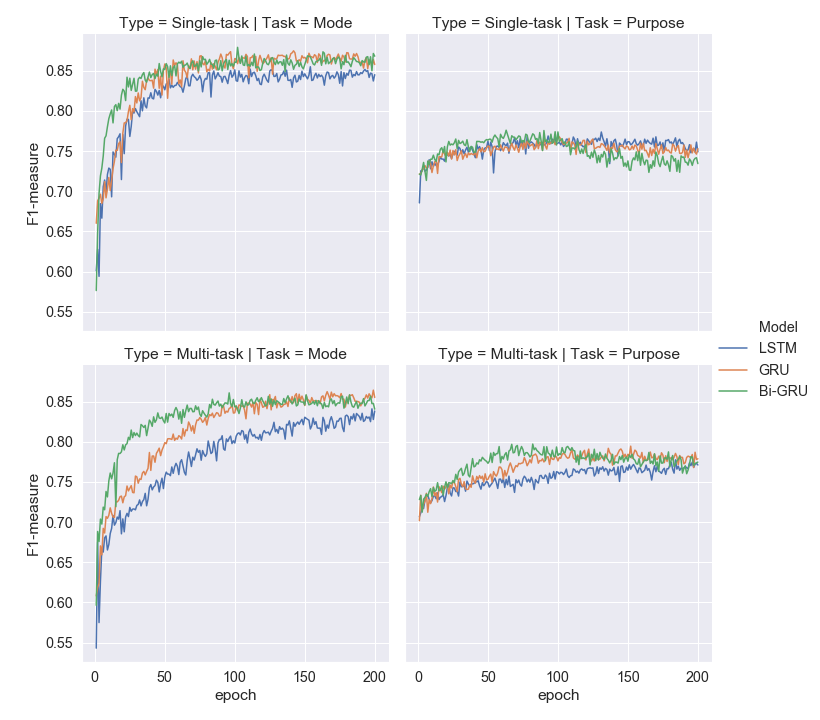}
\caption{F-1 Measure of Different Classifiers over Epochs (top row left: single-task mode, top row right: single-task purpose, bottom row left: multi-task mode, and bottom row right: multi-task purpose classifier)
\label{fig:total-results}}
\end{figure*}

All the models are trained to epoch 200, and the results are presented in Figure~\ref{fig:total-results}. Obviously, the performance of almost all the models in Figure~\ref{fig:total-results} do not improve beyond the 100 epochs. Hence, we reported and compared the performance of the models on the test data for epoch 100 in Table~\ref{Table:total-results}, in order to compare them on a fair ground. 

All models are implemented in Python, using the Keras backend API with GPU support. Different hyperparameter values were tested to select the best model architectures and configurations. Models are trained on the Google Colab supported GPUs with 12.0 GB of RAM. 

\begin{table*}[t]
\caption{Prediction Results of Different Purpose and Mode Classifiers (Epoch = 100, test data)}
\label{Table:total-results}
\resizebox{\textwidth}{!}{
\begin{tabular}{|c|c|c|c|c|c|c|}
\hline
Single/Multi Task                     & Task                                    & Model & Accuracy (\%) & Precision (\%) & Recall (\%) & F1-Measure (\%) \\ \hline
\multirow{6}{*}{Single-task Learners} & \multirow{3}{*}{Mode classification}    & LSTM         & 84.65         & 86.58          & 83.75       & 85.11           \\ \cline{3-7} 
                                      &                                         & GRU          & 86.07         & 87.49          & 85.58       & 86.50           \\ \cline{3-7} 
                                      &                                         & Bi-GRU       & 86.07         & 86.82          & 85.37       & 86.07           \\ \cline{2-7} 
                                      & \multirow{3}{*}{Purpose Classification} & LSTM         & 75.48         & 82.79          & 70.57       & 75.98           \\ \cline{3-7} 
                                      &                                         & GRU          & 75.99         & 83.62          & 71.02       & 76.59           \\ \cline{3-7} 
                                      &                                         & Bi-GRU       & 77.08         & 81.41          & 73.93       & 77.38           \\ \hline
\multirow{6}{*}{Multi-task Learners}  & \multirow{3}{*}{Mode classification}    & LSTM         & 79.91         & 83.50          &   76.67     &      79.83     \\ \cline{3-7} 
                                      &                                         & GRU          & 84.08         & 85.94          & 81.52       & 83.59           \\ \cline{3-7} 
                                      &                                         & Bi-GRU       & 84.08         & 86.24          & 82.61       & 84.33           \\ \cline{2-7} 
                                      & \multirow{3}{*}{Purpose Classification} & LSTM         & 76.70         & 84.67          & 69.79       & 76.20           \\ \cline{3-7} 
                                      &                                         & GRU          & 77.34         & 85.56          & 71.44       & 77.61           \\ \cline{3-7} 
                                      &                                         & Bi-GRU       & 77.92         & 84.41          & 73.34       & 78.28           \\ \hline
\end{tabular}}
\end{table*}

\subsection{Single-task Mode of Transport Classification}\label{single-task-purpose}
We developed three baseline single-task mode inference models to compare the performance of multi-task learners in comparison with them.  The accuracy, precision, recall and F1-measure of the single-task purpose classifiers are presented in Table~\ref{Table:total-results}. With respect to the F1-measure, the GRU model demonstrates the highest performance, with the F1-measure equal to 86.50\%. With respect to accuracy, the GRU and Bi-GRU models demonstrated equal accuracy of 86.07\%, and both were superior to the single-task LSTM mode classifier. Also, the GRU and Bi-GRU models achieved better recall values, 85.58\% and 85.37\% compared to the LSTM with recall equal to 83.75\%. Based on the results of the single-task mode classifiers in Table~\ref{Table:total-results}, the GRU models demonstrated a bit better performance compared to the LSTM and Bi-GRU models. 

\subsection{Single-task Trip Purpose Classification}
The LSTM, GRU and Bi-GRU models are developed testing their performance on single task trip purpose classification. Table~\ref{Table:total-results} demonstrates the accuracy, precision, recall and F1-measure of single-task purpose classifiers. The Bi-GRU model achieved the highest F1-measure of 77.38\%, while the LSTM and GRU models demonstrated F1-measures of 76.59\% and 75.98\%, respectively. Moreover, the Bi-GRU model is slightly better compared to both the GRU and LSTM with respect to recall. 

\subsection{Multi-task Mode and Purpose Classification}\label{sec:multi-task}
The multi-task learners developed use the same architecture as the single-task learners, except for the final loss functions, where two cross-entropy loss functions generate the final class labels. Optimization of the whole network is carried out by deploying an Adam optimizer on the sum of cross-entropy loss functions for mode and purpose.  

As shown in Table~\ref{Table:total-results}, the F1-measures for the multi-task LSTM learner for classifying mode and trip purpose are 79.83\% and 76.20\%, respectively. Comparing with the F1-measures of the single-task LSTM mode classifier (85.11\%) shows a considerable drop in the classification performance of the LSTM model when being deployed in a multi-task framework. However, the multi-task LSTM model predicted the purpose of trips with a F1-measure a bit higher than the single-task LSTM purpose classification. 

With respect to the multi-task GRU classifier, the model classifies the mode and purpose with F1-measure of 83.59\% and 77.61\%, respectively. Comparing these results with the single-task mode classifier shows that mode classification does not benefit from a multi-task framework, as the single-task GRU achieved an F1-measure of 86.50\%. Nonetheless, similar to the LSTM case, the multi-task GRU classifier depicts a higher F1-measure compared to the single-task GRU when classifying trip purpose. 

Regarding the multi-task Bi-GRU classifier, the model achieves an F1-measure of 84.33\%, which is higher than the F1-measure of both multi-task LSTM and GRU mode classifiers. At the same time, mode classification with the single-task Bi-GRU classifier shows superiority over the multi-task Bi-GRU classifier.

Based on the results in Table~\ref{Table:total-results}, the highest F1-measure for mode classification belongs to the single-task GRU model, while among all the purpose classification models the multi-task Bi-GRU achieved a bit higher F1-measure.

\subsection{Comparison with Other Learning Algorithms}
This section compares the performance of algorithms developed in the previous sections against other machine learning approaches: Random Forest (RF) and Convolution Neural Networks. Two RF models are developed to infer mode of transport and trip purpose. As the point-based GPS information cannot be fed to the RF models, the aggregated features of the whole trip have been used to predict mode of transport. The socio-demographics also have been used to enhance the prediction accuracy of the RF model. For trip purpose inference with the RF model, a single-task RF model is developed using land-use, Foursquare and socio-demographic data. The best RF model to predict the mode of transport achieves F1-measure of  85.68\%, while the RF model for trip purpose prediction reaches F1-measure of 71.53\%. The single-task GRU and Bi-GRU mode classifiers can predict the mode of transport better than the RF model, with F1-measures of 86.5\% and 86.07\%, respectively. However, the RF model shows better classification performance compared to all other mode classifiers in Table~\ref{Table:total-results}. Nevertheless, comparing the RF purpose classifier with the models in Table~\ref{Table:total-results} reveals the superiority of all single- and multi-task purpose classifier over the RF model. We also developed a Convolutional Neural Network (CNN) model, with the same characteristics as the best single mode CNN classifier suggested in ~\cite{yazdizadeh2019ensemble}, to detect mode of transport. The model can achieve F1-measure of 79.13\% to detect mode of transport.

\section{Conclusion}
Multi-task learning is assumed as a powerful inference method, specifically, where there is a considerable correlation between multiple tasks, predicting them in an unique framework may enhance prediction results. This research challenged this assumption by developing several single-task models to compare their results against multi-task learners to infer mode and purpose of trip from smartphone travel survey data. The comparison showed that, overall, mode and purpose inference does not benefit from multi-task learning approach. Indeed, based on the F1-measure, all the single-task mode inference models, slightly outperforms the multi-task learners. Also, regarding purpose of trip, the single-task LSTM and Bi-GRU showed a bit higher F1-measure over the similar multi-task learners, and only the multi-task GRU learner slightly outperforms the single-task GRU in predicting purpose of trip. In general, the results of the study does not hold the assumption of higher performance of multi-task over single-task learners in the context of mode and trip purpose inference from GPS trajectory data.

This study also examined the performance of well-known RNN architectures, i.e. LSTM, GRU, and Bi-GRU in the field of transportation data inference. The results demonstrated that for the single-task mode classifier, the GRU slightly outperforms the LSTM and Bi-GRU architecture. However, almost for all the other models (i.e. single-task purpose classifier and multi-task mode and purpose classifier) the Bi-GRU model showed higher performance over LSTM and GRU. 

Furthermore, this research used an entity embedding approach to encode categorical data, such as time of day and day of the week, to capture the correlation between different levels of each category. All the embedded layers were trained simultaneously with sequential GPS trajectory data and other numeric non-sequential data. However, while land-use and social network data around each trip destination are included in the models, the study lacks the inclusion of such information around GPS trajectories. That is, for example, infrastructure data, such as whether a metro line or highway or bike route is in the vicinity of a GPS trajectories, which may help the algorithm to detect mode of transport more accurately. Finding methods to include all such information around each GPS trajectories is left to future studies. 

Finally, besides the target tasks, such as mode and purpose in the current study, some studies \cite{augenstein2018multi} have proposed including auxiliary tasks in modeling procedures. Auxiliary tasks are those not as important as the target task for the researcher, but they may improve the performance of the inference model on target tasks. In the current study, traveler socio-demographics such as "occupation'' could play the role of auxiliary tasks instead of being fed as an input feature to the models. Examining such approaches may help the future studies to develop inference models, that achieve higher-performance. 

\section{Authors' Contribution}
Ali Yazdizadeh carried out the data processing, data analysis, and development of the machine learning algorithms. He also wrote the manuscript of the article. Arash Kalatian assisted in implementing the machine learning algorithms. Zachary Patterson and Bilal Farooq reviewed all of the methods and manuscript text. All authors discussed the results and contributed to the final manuscript.

\bibliographystyle{spbasic_updated}      
\bibliography{bib_library}   

\end{document}